# Robot Learning in the Era of Foundation Models: A Survey

Xuan Xiao[1,2,4], Jiahang Liu[1,2,3], Zhipeng Wang[1,2,3], Yanmin Zhou[1,2,3], Yong Qi[5], Qian Cheng[1,2,4], Bin He[1,2,3], Shuo Jiang[1,2,3]*

**Abstract:** The proliferation of Large Language Models (LLMs) has s fueled a shift in robot learning from automation towards general embodied Artificial Intelligence (AI). Adopting foundation models together with traditional learning methods to robot learning has increasingly gained recent interest research community and showed potential for real-life application. However, there are few literatures comprehensively reviewing the relatively new technologies combined with robotics. The purpose of this review is to systematically assess the state-of-the-art foundation model techniques in the robot learning and to identify future potential areas. Specifically, we first summarized the technical evolution of robot learning and identified the necessary preliminary preparations for foundation models including the simulators, datasets, foundation model framework. In addition, we focused on the following four mainstream areas of robot learning including manipulation, navigation, planning, and reasoning and demonstrated how the foundation model techniques can be adopted in the above scenarios. Furthermore, critical issues which are neglected in the current literatures including robot hardware and software decoupling, dynamic data, generalization performance with the presence of human, etc. were discussed. This review highlights the state-of-the-art progress of foundation models in robot learning and future research should focus on multimodal interaction especially dynamics data, exclusive foundation models for robots, and AI alignment, etc.

**Keywords:** Robot Learning, Foundation Models, Embodied AI

## 1 Introduction

Robots have played an important role in various scenarios including industrial[1], medical[2], service[3] and special robot[4] industry. With the increasing task complexity and working environment variability, the demands for robot tasks have shifted from fixed automation to general artificial intelligence, where robot learning will be the core enabling techniques of the autonomous systems[5].

The field of robot learning lies at the junction of machine learning and robotics. Its primary focus is enabling robots to acquire new skills or adapt to their environment by utilizing learning algorithms. Some examples of skills that learning algorithms target include sensorimotor skills (like locomotion, grasping and active object classification), interaction skills (such as co-manipulating objects with humans) and linguistic skills (including the grounded and situated meaning of human language). Learning can be acquired through autonomous self-exploration or through demonstrated teaching.

Extensive research studies have been performed on how to realize robot learning. Traditional robot learning techniques are generally divided into imitation learning[6-9] and reinforcement learning[10, 11]. Imitation learning enables robots to learn new skills by mimicking human behavior, while reinforcement learning allows robots to optimize the outcome of skill execution. Imitation learning is an important way to initialize and improve learning efficiency in reinforcement learning. However, they all suffer from certain limitations. Imitation learning is a straightforward and stable form of supervised learning. It requires labeled behavior data and is unable to surpass human-level performance. Although reinforcement learning has the potential to surpass human-level performance, it requires defining the reward function, addressing the policy exploration challenge, and may encounter convergence issues and instability during training. The characteristics of robot learning is that robots are physically embodied and environmentally situated. Thus, a critical challenge in robot learning is to close the perception-action loop in the practical scenarios and there are still some problems in the practical application of traditional robot learning, including insufficient generalization of tasks, insufficient environmental adaptability, low execution accuracy, and lack of planning and reasoning capabilities[12-14].

Recently, with the release of ChatGPT, foundation models, particularly multi-modal foundation models, presented both opportunities and challenges for robot learning. LLMs have demonstrated significant potential in achieving human-level intelligence, thereby leading to a surge of research in robotics based on LLMs. Leveraging multi-domain prior knowledge, LLMs have made breakthroughs in understanding complex task, engaging in

---

[1] National Key Laboratory of Autonomous Intelligent Unmanned Systems, Tongji University, Shanghai, China.

[2] Frontiers Science Center for Intelligent Autonomous Systems, Ministry of Education, Tongji University, Shanghai, China.

[3] College of Electronics and Information Engineering Tongji University, Shanghai, China.

[4] Institute of Acoustics, School of Physics Science and Engineering, Tongji University, Shanghai, China.

[5] School of Electronic Information and Artificial Intelligence, Shaanxi University of Science and Technology, Shaanxi, China.

This work was supported by National Key Research and Development Program of China under Grant 2020AAA0108905, National Natural Science Foundation of China (Grant No. 62088101, 52105033), and the Shanghai Municipal Science and Technology Major Project (2021SHZDZX0100). (*Corresponding author：Shuo Jiang) continuous dialogue, and performing zero-sample reasoning. However, LLMs still lack the general perception ability of the external environment, which can be addressed by employing the multi-modal foundation model incorporating 2D&3D vision, LiDAR, voice, inertial motion unit (IMU), etc. To fully unleash the potential of the foundational model and address the current challenges in robot learning, so that robots can learn human behavior and skillfully undertake a series of tasks, researchers have developed task-specific robot learning architectures. However, these models were proposed independently and relatively recently. Upon thorough investigation, existing surveys on robot learning are primarily focused on a single task and predominantly rely on traditional methods[5, 15-19]. There are few literature reviews on multi-task robot learning based on foundation models. Therefore, it is crucial to construct an overall summary analysis of existing LLM-based robot learning work, which is of great significance for a comprehensive understanding of the field and to provide inspiration for future research[20].

We organize this survey according to the solutions of the following three research questions. What platforms does the embodied AI need? What foundation model algorithms, strategies, and mechanisms are currently used for downstream tasks in robot learning? What are promising future research areas in robotics learning for current research questions? In this paper, we systematically survey the relatively new field of foundation model-based robot learning, establish a clear taxonomy for the existing research in the field, focusing on the four aspects of robot learning: manipulation, navigation, planning, and reasoning. We also identify several challenges in this field and discuss potential future directions. The article follows the organizational structure presented below (Fig. 1).

We select references based on the following criteria: 1) Our primary objective is to encompass all significant milestones in robot learning during the advent of foundation models across diverse tasks. 2) the diversity and significance of subfields and research groups should be considered. 3) A valuable source entails conducting a reverse search beginning with highly influential publications[21].

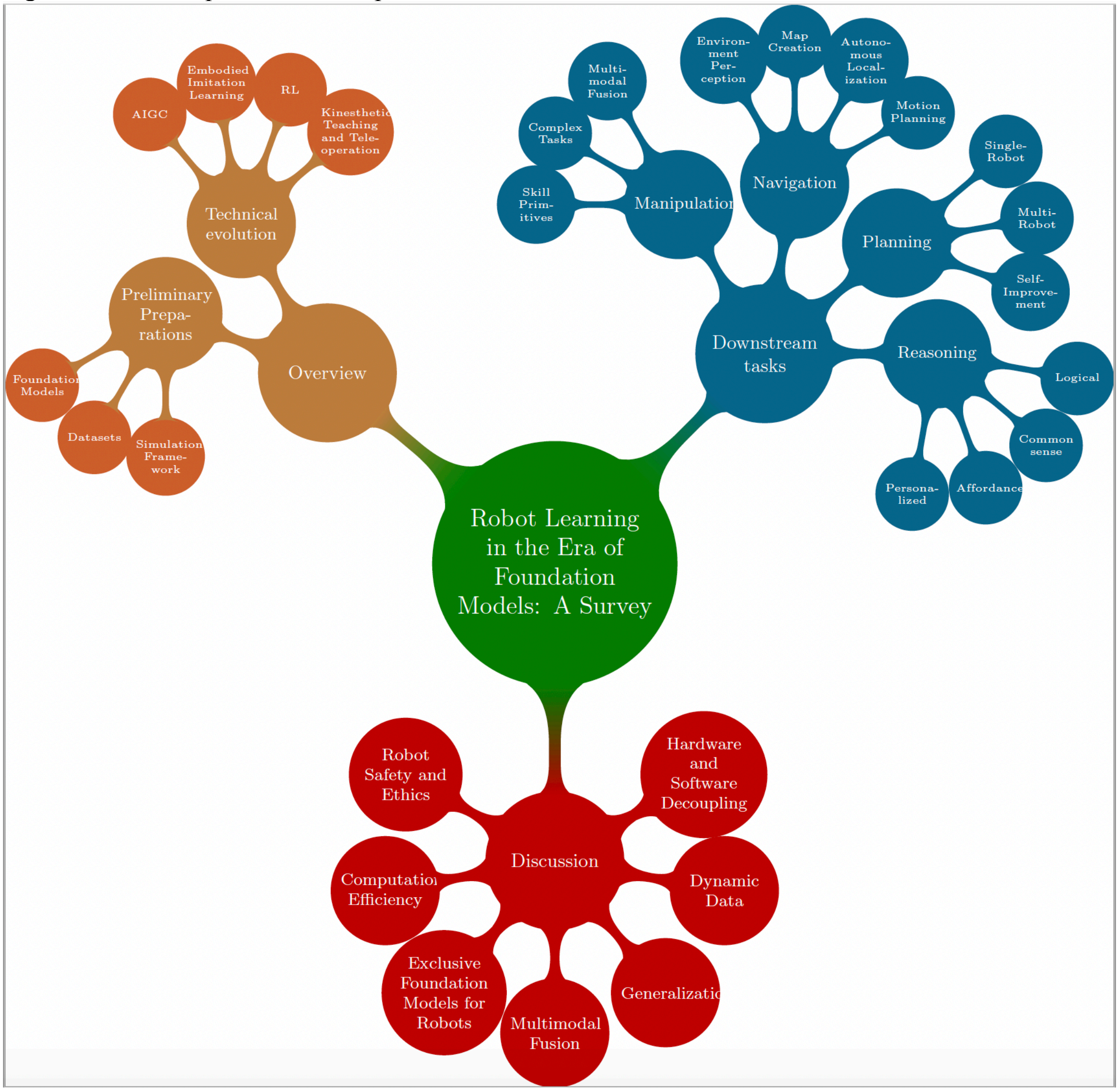

Fig.1. Overall structure of the survey.

## 2 Overview

### 2.1 Technical Evolution

In the exploration and discovery of robot learning, it has gone through different stages of historical evolution (Fig.2). They can be summarized as the following four categories: robot learning through teaching programming (TP) including kinesthetic teaching and teleoperation, reinforcement learning in interaction, embodied imitation learning [22, 23] and learning in AIGC-based generative models[24-26].

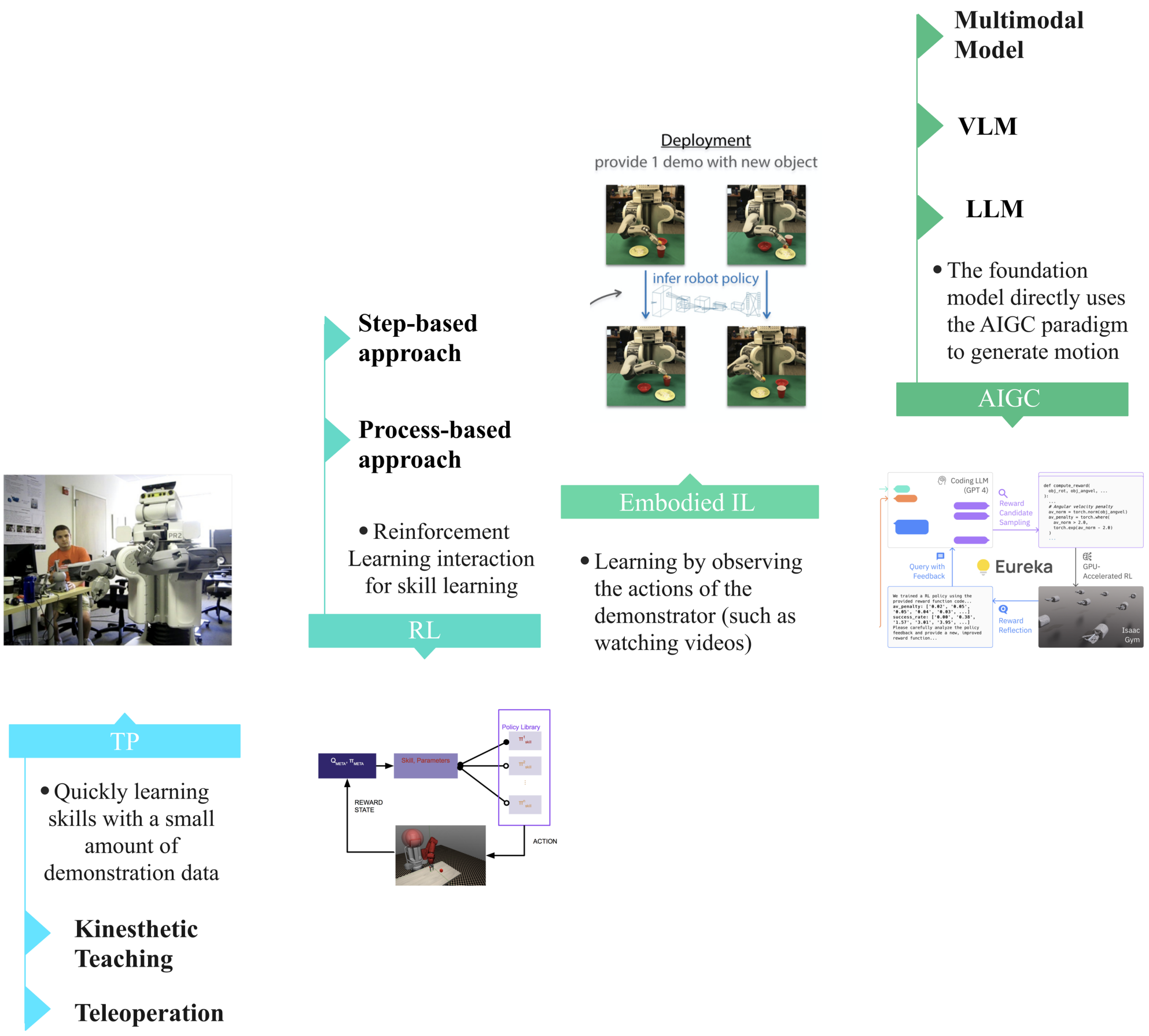

Fig.2. Technical Evolution[27-30].

Kinesthetic teaching[31] involves physically maneuvering the robot to perform the necessary actions, while the robot's onboard sensors record status information, which is then used to generate training data for the machine learning model. Although this method is relatively simple, the quality of the demonstration relies heavily on the operator's ability to perform movements with flexibility and smoothness. While it proves highly effective for robotic arms, its application becomes more challenging on other platforms, such as legged robots or dexterous hands. Another demonstration method is teleoperation[32], which enables trajectory learning, task learning, grabbing or more advanced tasks, by providing external input to the robot through handles, graphical interfaces or other methods. There are currently a variety of interactive devices (such as haptic devices or VR interactive devices). In contrast to kinesthetic teaching, teleoperation allows for remote implementation, eliminating the need for the user and the robot to be physically present at the same site. Limitations of teleoperation include the additional work required to develop input interfaces, a longer user training process, and usability risks posed by external devices.

Reinforcement learning involves determining optimal actions in a given situation to maximize the digital revenue signal. Rather than being explicitly instructed on which actions to take, learners must independently identify the actions that yield the greatest benefits. Reinforcement learning focuses on finding a balance between exploring unknown territory and leveraging current knowledge. A typical framework for a reinforcement learning scenario: an agent takes an action in the environment, and the action is interpreted as a reward and state representation, which is then fed back to the agent [33].

Embodied imitation learning[34] mainly involves robots learning by observing the actions of the demonstrator (such as watching videos). In this method, the presenter utilizes their own body to perform the task, while an external device records their movements. This approach proves to be the most straightforward for presenters, as it does not necessitate any training in the presentation process. Furthermore, this approach can be extended to robots with multiple degrees of freedom, including non-anthropomorphic robots. However, enabling this approach involves mapping human actions to robot-executable actions, which presents challenges such as occlusion, rapid movement, and sensor noise during demonstrations.

By integrating foundation models[25] with robots, based on their embodied intelligence, zero-sample training is achieved for motion planning and execution in real-world scenarios, resulting in human-like motion control logic and capabilities. This type of robot obtains general knowledge from foundation models and addresses cognitive deficiencies through perceptual intelligence, such as vision. This marks the initial steps towards the industrialization of embodied intelligent robots.

## 2.2 Preliminary Preparations

Developing a learning model for robots is not an easy task[35, 36], given the challenges of technical issues[37, 38] including computing resources, algorithms and data. A feasible approach is to perform incremental development or experimental verification on the basis of existing models. In this section, we briefly organize publicly available resources[39-45] for robot learning, including simulators, datasets[46-49], foundation models[36].

As developing and testing applications with real robots is expensive and time-consuming, simulation has emerged as a crucial component in the field of robotics application development. The validation of applications in simulation prior to deployment on robots can shorten iteration time by identifying potential issues at an early stage. Simulation facilitates the testing of corner cases or scenarios that could pose risks in the real world. To validate and assess the performance and effectiveness of robot learning, it is beneficial to conduct initial testing in a simulator before transitioning to the real-world environment. Several simulation frameworks[50] currently exist for training in robot learning simulations and are summarized in Appendix Table 1.

To validate the correlation performance in the simulation experiment, it is essential to have a dataset. There are two types of data sets: static data and dynamic data. Static data refers to related data sets collected on the Internet, and dynamic data refers to data generated through interactions with robots in a real-world environment. Currently, the majority of existing datasets consist of static data, and the specifics of prominent datasets are summarized in Appendix Table 2. Obtaining dynamic data is challenging, and it is relatively scarce.

Foundation models[51] are inseparable from the support of computing power, algorithms and data. The chip determines the computing power. Higher-performance chips are necessary for the training and construction of the neural network in foundation models. Research institutes currently employ diverse algorithms to implement foundation models. However, the primary difficulty lies in acquiring high-quality data. High-quality data plays a crucial role in facilitating AI training and tuning. The following is a summary of notable foundation models in Appendix Table 3.

Due to the substantial cost of model pre-training, utilizing public API[52, 53] allows for remote execution of inference tasks.

# 3 Downstream Tasks

The target and output of robot learning is to help the robot tactfully fulfill a certain task via acquiring a behavior or skill. Common target tasks include robot manipulation, navigation, mission planning, and reasoning (Fig.3). This section investigates and reviews how different works combine foundation models for different downstream tasks to achieve robot learning goals[54, 55].

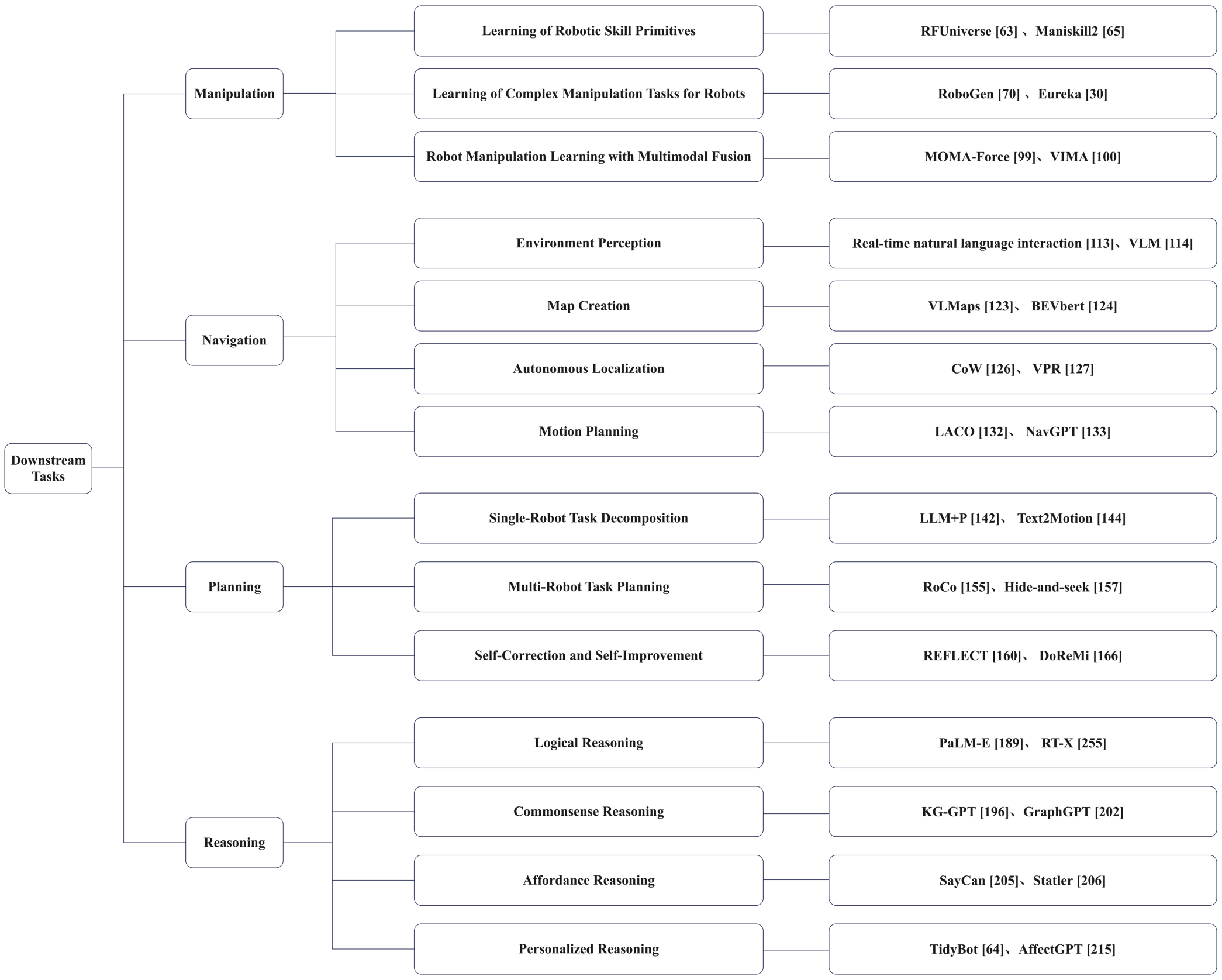

Fig.3. Downstream Tasks For Robot Learning.

### 3.1 Manipulation

#### 3.1.1 Characteristic and Challenges

Tasks, despite being commonly perceived as simple and routine, such as washing dishes, cutting vegetables, and packing clothes, are still challenging for robots and are at the forefront of robotics research[56]. The robot manipulation problem deals with how a robot learns to manipulate its surrounding environment[5]. It can be formulated as the ability to connect a starting state to a goal state through successive actions, during which the manipulation task is represented by a set of points denoting the starting and goal states, along with the constraints imposed on the transition states[57]. Furthermore, the ability to make physical contact is critical to the development of manipulation skills. Currently, robots can only effectively grab and release certain types of objects and perform a variety of simple manipulation actions such as throwing, sliding, pushing, and poking. Challenges emerge when these actions need to be performed in uncluttered environments or when more complex interactions are necessary[14]. Foundation models have the capability to engage in interactive dialogues with users, receive and process various types of data, including images, text, and speech. They can utilize multimodal information, such as vision, to guide the actions of the robot or generate corresponding code or action sequences. These capabilities enable robots to effectively adapt to intricate and ever-changing environments, eliminating the need for defining every single possible scenario in advance[55]. In this section, we survey the recent research results and latest progress of foundation models in the direction of robot manipulation, including learning of robotic skill primitives, learning of complex robot manipulation tasks, and robot manipulation learning with multimodal fusion.

#### 3.1.2 Methods

- Learning of Robotic Skill Primitives

The approach involves breaking down the problem of robot skill learning into several easily comprehensible skill primitives or movement primitives. Subsequently, suitable and efficient learning methods are devised for these skill primitives, allowing the robot to acquire manipulation skills by combining these primitives. This enables the robot to adapt to new environments and generalize its abilities to perform novel tasks. Researchers are currently exploring the utilization of LLM guidance to facilitate the learning of action primitives for skill completion[58, 59].

The ActivityPrograms knowledge base[60] collected by Puig et al. has 292 different high-level tasks in the knowledge base. This work lays the foundation for further exploration by future generations. Huang et al.[61] randomly sampled 88 holdout tasks from the aforementioned knowledge base for evaluation, and the remaining 204 tasks were used as a demonstration set, from which they were used as examples for prompting language models. For supervised fine-tuning baselines, these tasks were utilized in fine-tuning the pretrained language model. Brohan et al.[62] collected a large and diverse dataset of robot trajectories, including multiple tasks, objects, and environments. The main dataset contains 130,000 robot demonstrations, which were performed by 13 robots over a 17-month period. The robot conducted large-scale data collection in a series of office and kitchen. Fu et al.[63] proposed a novel physics-based action-centric environment, RFUniverse, for robot learning of daily housework tasks. RFUniverse supports 87 atomic operations and interactions between 8 primitive object types in a visually and physically plausible way. To demonstrate the usability of the simulated environment, the learned algorithm is performed on various types of tasks, namely fruit picking for manipulation, folding cloth and sponge wiping, stair chasing for locomotion, room cleaning for multi-agent collaboration, milk pouring, and hand lift for cloning behavior from VR interface. Vaucher et al.[22] extracted 28 kinds of atomic operations for synthetic operations of chemical experiments using three models, which are rule-based action extraction from Pistachio, rule-based action extraction from NLP, and a combination of the above two methods. Wu et al.[64] propose to use pick-and-place and pick-and-throw as action primitives, arguing that they are well suited for household cleaning, and thus propose a personalized robot assistance model with a large language model[58]. Gu et al. introduced the benchmark maniskill2, comprising a series of 20 operational tasks specifically designed to tackle the major challenges faced by researchers when utilizing general operational skill benchmarks[65]. Gao et al. proposed a two-stage fine-tuning strategy to further enhance the model's generalization ability based on the maniskill2 benchmark[59]. Furthermore, there are additional pertinent works that can be referenced[66-69].

- Learning of Complex Manipulation Tasks for Robots

With the expanding scope of application scenarios and increasing performance requirements for robots, the demand for completing increasingly complex and precise manipulation tasks has risen. Therefore, the research goal is to empower robots with the capability to autonomously learn from their environment and independently accomplish complex manipulation tasks. This includes acquiring knowledge, accumulating experience, continuously updating and expanding manipulation skills in order to successfully execute complex tasks.

Wang et al. proposed RoboGen, a method for achieving robot automation learning by generating simulated infinite data[70]. Ma et al. proposed the Eureka method, which utilizes LLMs for human-like reward design. They demonstrated, for the first time, a simulated shadow hand capable of performing pen spinning techniques[30]. Shridhar et al. propose CLIPORT, a language-conditioned imitation learning agent that combines the broad semantic understanding of CLIP with the spatial precision of Transporter, an end-to-end framework capable of solving various language-specified desktop tasks[71]. Khandelwal et al. build a simple baseline EmbCLIP without task-specific architectures, inductive biases (e.g. using semantic maps), auxiliary tasks during training, or depth maps, but the improved baseline performs remarkably well across a range of tasks and simulator[72]. Wang et al. access pre-trained visual language models in robot manipulation, using a combinatorial classification grammar that parses sentences into operational procedures in a domain-specific language, which allows visual language models to correspond category or attribute descriptions to pixels, thus opening up the learning of visual and action strategies[73]. Xiao et al. introduced Data-Driven Instructional Augmentation with Linguistic Condition Control (DIAL) and generalized it to 60 new instructions not seen in the original dataset. This effectively incorporates internet-scale knowledge into existing datasets with limited real annotations[69]. Huang et al. used LLMs to infer affordances and constraints through language, used its code writing ability to interact with VLM to generate 3D value maps, integrated knowledge into observation space, and used value maps to model zero-shot synthesis of closed-loop robot trajectories. Robust to perturbations, the model is used to perform a variety of manipulation tasks[74]. Mo et al. proposed SeeAsk, an open-world interactive vision-based system that can master specific goals through vague natural language instructions[75]. Ze et al. proposed GNFactor, a visual behavioral cloning agent for multi-task robot manipulation[76]. Zheng et al. designed the canonicalized manipulation space and proposed a two-stage framework for synthesizing human-like manipulation animations covering rigid and articulated object categories[77]. Extensive research has also been conducted on dexterous manipulation [78-96].

- Robot Manipulation Learning with Multimodal Fusion

While multimodality[97] addresses the challenge of agent perception, it falls short in addressing the issue of cognition. In various scenarios like an intelligent customer service, users often provide information through multiple modalities. Multimodality holds significant perceived value, but it still has substantial challenges to overcome in solving fundamental issues. Multimodality represents a prominent future trend wherein foundation models will progressively align with multimodal approaches, enabling future agents to operate within a multimodal environment[98].

Yang et al. proposed MOMA-Force, an imitation method for visual binding force, using perceptual representation learning, imitation learning, and admittance whole-body control to enhance the robustness and controllability of the system. MOMA-Force enables mobile manipulators to learn multiple complex contact-rich tasks with high success rate and small contact force[99]. Jiang et al. demonstrate that a wide range of robotic manipulation tasks can be expressed with multimodal prompts, and many robotic manipulation tasks can be represented as multimodal prompts interleaved with language and images or video frames. The authors design a transformer-based robotic agent, VIMA, that processes these prompts and outputs motor actions autoregressively[100]. Sferrazza et al. achieve robot generalization manipulation through multimodal learning that fuses vision and touch. AudioPaLM combines text-based and speech-based language models[101]. Brohan et al. propose to jointly fine-tune state-of-the-art visual-language models on robot trajectory data and Internet-scale visual-language tasks. The authors present RT-2: Vision-Language-Action Model Transferring Network Knowledge to Robot Control[102]. RobotGPT provides the perception, cognition, decision, execution capabilities with embodied AI[103]. Luo et al. proposed a physics-based humanoid controller that enables high-fidelity motion imitation and fault-tolerant behavior in the presence of noisy inputs (e.g., pose estimates generated from videos or speech) and unexpected falls[104]. Gandhi et al. conducted the first large-scale study of the interaction between sound and robot movement. The authors created the largest available sound-action-vision dataset using the robotics platform Tilt-Bot, with 15,000 interactions on 60 objects[105]. Peng introduces Kosmos-2, a multimodal large language model (MLLM) that supports perceptual object descriptions (such as bounding boxes) and new capabilities for integrating text into the visual world[106]. Radosavovic et al. proposed an RPT model, which is a Transformer that operates on a sequence of sensorimotor tokens. Given a sequence of camera images, proprioceptive robot states, and past actions, it is assumed that if the robot can predict what is missing, then it has acquired a good model of the physical world that allows it to act[107]. Li et al. systematically studied how visual, auditory, and tactile perception can jointly help robots solve complex operating tasks[108]. LEE et al. use graph neural networks to generate master odor maps (POMs) that preserve perceptual relationships and enable odor quality predictions for previously uncharacterized odors[109]. Guzey et al. proposed visually stimulated tactile adaptation in order to learn tactile flexibility, thereby correcting errors and adapting to changing situations[110].

### 3.1.3 Datasets and Metrics

**Datasets.** In order to improve the generalization ability of robot manipulation learning, researchers collected multi-task, multi-scenario data sets so that the robot can be trained to quickly generalize various scenarios. At the same time, the test datasets evaluate the manipulation capabilities of these robots guided by the foundation model. There are several datasets available for testing, as presented in Appendix Table 4.

**Metrics**. In order to improve robot performance and manipulation level, the key is to continuously optimize the performance of the robot learning system. Then the core issue is to reasonably evaluate the performance of the robot. Issues encountered in the comprehensive evaluation process: improvement of single evaluation indicators, integration of multiple evaluation indicators, weight distribution of each indicator, and reliability of evaluation results.

Manipulation performance is evaluated by: (1) Plan success rate, which measures whether the skills selected by the model are correct for the instructions, regardless of whether they are actually successfully executed. (2) Execution success rate, which measures whether the entire system actually successfully executed the required instructions[62, 102].

### 3.1.4 Problems

Despite making considerable progress, significant problems still exist. Robot movements are continuous and complex, with more physical interactions and manipulation causality.

A comprehensive library of skills and actions needs to be built. In order to complete downstream target tasks, a skill library needs to be built. High-level instructions are broken down into atomic actions. The existing skill action library suffers from a long-tail distribution problem. A more comprehensive skill action library needs to be established to assist robot skill learning. Robots can manipulate objects of any size, shape, and material in any scene. For example, a robot can pour a specified volume of any liquid and complete specified tasks in a dynamically changing environment. Objects that are difficult to manipulate in the 3D world, such as articulated, deformable and transparent objects, can be manipulated. The robot can solve the problems of occlusion and obstacle avoidance when grasping. The robot can reason about spatial relationships and contact dynamics, accurately drive high-degree-of-freedom arms, and apply the appropriate force to stably grasp objects without breaking them.

Current robotic object manipulation applications rely heavily on human programming and have poor autonomy, which is unrealistic in the long run. Vision-based intuitive physics, open set grasping with semantic representation, and planning under partial observability and uncertainty will be the future trends of robot grasping[15].

### 3.2 Navigation

#### 3.2.1 Characteristic and Challenges

Several research projects focus on developing autonomous robots capable of navigating in different environments. However, little research has specifically addressed the navigation challenges faced by legged robots in indoor environments, especially when relying on single-camera vision. The legged robot is able to traverse uneven surfaces and overcome obstacles, such as stairs, that traditional wheeled robots typically cannot reach. In order to realize the autonomous navigation of mobile robots in unknown dynamic environments, mobile robots need to solve three basic problems, namely "Where am I?", "What is my surrounding?", and "What should I do next?". In order to solve these three core basic problems, a series of core technologies such as environmental perception, map creation, autonomous positioning, and motion planning are required[111, 112]. Foundation models have reasoning and decision-making capabilities, and multi-modal foundation models can handle rich information modalities, and encode information of different modalities into the same vector space, which is conducive to cross-modal information processing. Benefiting from these considerations, foundation models help to deal with problems such as motion planning and environment perception of robots. In this section, we survey recent research results and state-of-the-art advances in the direction of robotic navigation with foundation models, including robotic environment perception, map creation, autonomous localization, and motion planning.

#### 3.2.2 Methods

- Environment Perception

Environmental perception refers to the process in which a robot, during its movement, perceives the surrounding environment through various devices. It can collect real-time environmental information and adjust its position and pose accordingly based on the actual situation.

Lynch et al. proposed a robot framework for real-time natural language interaction in a real environment. The relevant resources are open source (datasets, environments, benchmarks, and policies). Behavioral cloning training is performed on the dataset of annotated trajectories, and the generated Policies proficiently execute orders of magnitude more commands than previous work[113]. Methods that align the embedding space of different modalities (in this case, IMU data) with the visual embedding space enable the VLM to understand and reason about these additional modalities without the need for retraining. The results show that using multiple modalities as input can improve VLM's scene understanding and enhance its overall performance in various tasks[114]. The agent follows real-life navigation instructions, then recognizes the location described in natural language and finds the object at the target location[115]. Hong et al. proposed a predictor to generate a set of candidate waypoints during navigation, bridging the gap between learning in discrete and continuous environments for visual and language navigation[116]. Tan et al. use environment dropout for back-translation to learn unseen environments in navigation[117]. Qi et al. propose an object and action perception model that can flexibly match object-centered or action-centered instructions with their corresponding visual perception or action direction in the process[118]. Various other studies can serve as references[119-121].

- Map Creation

Map construction can be accomplished either through the utilization of metric maps or by employing symbols that represent the robot's position in its reference frame.

Gervet et al. proposed a semantic visual navigation method, compared six typical approaches of classical, modular and end-to-end learning methods without any experience, maps or instruments, through large-scale empirical research, found that modular learning is a reliable way to navigate to objects[122]. To address the lack of spatial accuracy of classical geometric maps, Huang et al. proposed spatial map representation VLMaps, which fuses pretrained visual-linguistic features with 3D reconstructions of the physical world. VLMaps can be autonomously constructed from the robot's video sources using standard exploration methods and can support natural language indexing of maps without labeled data[123]. An et al. proposed a multi-modal map pre-training paradigm BEVbert for language-guided navigation, which has spatial awareness capabilities[124]. Jia et al. proposed a neural SLAM method that for the first time exploits multiple modes to explore, predict perceptible semantic maps, and plan them simultaneously[125]. Zhang et al. propose a hierarchical object-to-region (HOZ) environment map to guide the agent in a coarse-to-fine manner. Additionally, an online learning method is proposed to update the HOZ new environment based on real-time observations.

- Autonomous Localization

Robot localization refers to a robot's ability to accurately determine its position and orientation within a predefined reference frame. Inspired by the recent success of open-vocabulary models for image classification, Gadre et al. investigate a simple framework, CLIP on Wheels (CoW), to adapt open-vocabulary models to language-driven zero-shot object navigation without fine-tuning (L-ZSON)[126]. Keetha et al. have developed a universal VPR solution - a technology that works across a variety of structured and unstructured environments without the need for any retraining or fine-tuning[127]. Based on the WAY dialogue dataset, Hahn et al. focus on the LED task (locating the observer from the dialogue history)[128]. Two agents (tourist and tour guide) interact through natural language to allow tourists to navigate to the correct location[129].

- Motion Planning

Motion planning is the process by which a robot intelligently selects an optimal sequence of actions to move towards a target location based on its current position within a reference frame, considering changes in the environment. Hong et al. proposed a time-aware recurrent bert model to maintain cross-modal state information in order to solve the problem that the bert architecture in visual language navigation is difficult to use for partially observable Markov decision-making processes, which need to rely on historical attention and decision-making[130]. Manglani studies the development of visual segmentation and path planning algorithms for autonomous navigation and obstacle avoidance systems in domestic environments[111]. Lin et al. propose to provide action prompts for VLN agents to enable explicit learning of action-level modality alignment for navigation. Specifically, action prompts are defined as a pair of modality-aligned image sub-prompts and text sub-prompts, and when navigation starts, a set of action prompts related to an instruction is retrieved from a pre-built library of action prompts and passed through a prompt encoder to get hint features. Then, the prompt features are concatenated with raw instruction features and fed to a multi-layer transformer for action prediction. Using a contrastive language-image pre-training (CLIP) model, a modality alignment loss and a sequential consistency loss are introduced to enhance the alignment of action prompts and force the agent to sequentially attend to related prompts[131]. Traditional path planning algorithms focus only on collision-free paths, limiting their applicability in contact-rich tasks. To address this limitation, Xie et al. proposed Linguistically Conditional Collision Function (LACO), a novel method to learn collision functions using only single-view images, language cues, and robot configurations[132]. Zhou et al. introduce NavGPT, an LLM-based instruction tracking navigation agent, to demonstrate the reasoning ability of large language models in complex scenes by performing zero-shot sequential action prediction for visual and language navigation[133]. Qian et al. propose the March-In-Chat model, which allows communication with LLMs in real time and perform dynamic planning based on room-and - object aware scene perceiver. The model outperforms previous models on the REVERIE benchmark[134]. Fried et al. experimentally demonstrated that the three parts of the proposed method (speaker-driven data augmentation, pragmatic reasoning, and panoramic action space) greatly improved the performance of the baseline instruction follower[135]. Wang et al. proposed enhanced cross-modal matching and self-supervised imitation learning methods to solve cross-modal grounding, ill-posed feedback and generalization problems[136]. Nguyen et al.[137] develop a memory-enhanced neural agent. Imitation learning algorithms teach agents to avoid repeating past mistakes. The agent also completes object finding tasks by requesting and interpreting natural language and visual assistance.

#### 3.2.3 Datasets and Metrics

**Datasets.** In order to promote the development of robot navigation tasks, researchers have collected a large amount of data from multiple aspects such as instructions, scenes, objects, etc. This plays a very important role in the development of navigation and also lays the foundation for the transfer from virtual to real environments. There are many datasets that can be used for navigation. See Appendix Table 5 below.

**Metrics.** Navigation performance is evaluated by: (1) Success / Oracle Success Rate (%). (2) Navigation Error (m). (3) SPL (Success weighted by Path Length). (4) CLS (Coverage weighted by Length Score), measuring fidelity to the reference path. (5) nDTW (normalized Dynamic Time). (6) SDTW (Success weighted by normalized Dynamic Time Warping)[138-140].

#### 3.2.4 Problems

To solve the problem of high obstacles, it is worth considering investigating partially observable search methods. Potential solutions would involve broadening the scope of the study to cover a wider range of domestic environments (including stairs and uneven surfaces), outdoor environments and complex unknown environments. In addition, a more general large-scale navigation model can be designed that can be widely applied to any robot, similar to LLMs and VLMs for processing any text or image.

In addition, the efficiency of multi-target navigation needs to be improved, scene knowledge needs to be better applied in navigation, and the efficiency of unknown scene navigation needs to be improved. Techniques for optimizing system efficiency in real-time scenarios will also be explored, taking into account factors such as robot stability, dynamic obstacle avoidance, and resource constraints.

Path planning in unknown scenarios faces the following problems: lack of global map, generalization of unknown environments, and sensor noise in real-life applications. The lack of global map includes the following aspects: lack of global SLAM map, how to learn prior knowledge of layout, and in what form to store knowledge. The generalization of unknown environments includes the following aspects: limited training rooms and infinite unknown environments, transfer and generalization of knowledge. Sensor noise in real-life applications includes the following aspects: depth loss, RGB blur, and detection and segmentation errors.

### 3.3 Task Planning

#### 3.3.1 Characteristic and Challenges

Depending on the difficulty of the task, the task can be completed by a single robot or by multiple robots. Many of these problems require multiple robots to work together in multi-robot systems. Long-term tasks for a single robot need to be broken down into smaller tasks that can be accomplished with simple operations. Task planning for multiple robots working together needs to consider the relationship between tasks, robot capabilities and cooperation, and other challenges. In addition, the agent cannot hallucinate when planning tasks. Within the bounds of common sense, agents can learn from previous tasks, self-correct and improve, and continue to learn. This is a challenging task. The true realization of intelligent robots will be based on the perception of the environment, dynamic learning, and continuous updating. In this section, we survey the recent research results and the latest progress of foundation models in the direction of robot task planning, including single-robot task decomposition, multi-robot task planning, and self-correction, self-improvement, and continuous learning of agents.

#### 3.3.2 Methods

- Single-Robot Task Decomposition

Task planning. For complex tasks, the best solution is to split it into several simple tasks, and then solve them one by one. Generally, the effect will be good. LLMs have shown good zero-shot generalization ability. State-of-the-art chatbots can provide plausible answers to many common questions that arise in everyday life. However, so far, LLMs cannot reliably solve long-term planning problems[141].

Liu et al. introduce LLM+P, the first framework to incorporate the strengths of the classical planner into LLM[142]. Wu et al. aligned the LLM with the visual perception model, and generated executable action sequences based on the objects existing in the real scene[143]. Lin et al. proposed a language-based planning framework, Text2Motion, based on which robots can solve sequential manipulation tasks for long-horizon reasoning. This framework uses the feasibility heuristic encoded in the skill library Q function to guide the task planning of LLMs, and Geometric dependencies among skills are resolved by performing geometric feasibility planning during the search process[144]. Wake et al. use ChatGPT to convert natural language instructions into executable robotic actions in long-step scenarios[145]. EmbodiedGPT introduces an efficient training method to generate high-quality plans. The paradigm of extracting task-relevant features from LLM-generated planning queries is introduced to form a closed loop between high-level planning and low-level control[146]. Ruan et al. proposed a structured framework customized for LLM-based artificial intelligence agents, within which two different types of agents were designed: one-step agents and sequential agents, to perform the reasoning process. The framework is then instantiated using LLMs and evaluated on their task planning and tool usage abilities on typical tasks[147]. Zhen et al. propose a task planning method that combines human expertise with LLMs. And the LLM prompt template is designed, which has stronger expressive ability to represent structured professional knowledge. Further, a method of gradually decomposing tasks to generate task trees is proposed. And a strategy for decoupling robot task planning is designed[148]. Gu et al. studied a modular approach to handle the long-view movement operation task for object rearrangement, decomposing the complete task into a series of subtasks[149]. Obinata et al. describe a strategy for implementing a robotic system capable of performing general service robot (GPSR) tasks in robocup@home[150]. Shi et al. developed RoboCook, an intelligent robotic system that can use multiple tools for long-horizon elastoplastic object manipulation[151]. There are some other works that can be referenced[61].

- Multi-Robot Task Planning

For multiple robots, relationships between tasks needs to be considered during task planning, robot capabilities and cooperation, and other challenges. Task planning for multi-robot teams usually involves three sub-problems: task decomposition, task allocation and task scheduling, which are introduced here to consider situations where robots sequence their own tasks to satisfy various constraints. While each of these problems can be solved individually, the interdependencies between these subproblems are often considered in the context of integrated tasks to provide solutions that are often better than solving each subproblem independently[18, 152, 153].

LLMs have shown impressive planning capabilities in single-agent specific tasks across domains, yet their planning and communication capabilities in multi-agent cooperation remain unclear. A novel framework for multi-agent cooperation using LLMs is proposed, which enables agents to plan, communicate and cooperate with other agents or humans to effectively complete long-horizon tasks[154]. RoCo is a unified approach for multi-robot collaboration that leverages pre-trained LLMs for high-level communication and low-level path planning. The authors introduce the RoCoBench benchmark, which includes various challenges in collaborative task scenarios. The method demonstrates its utility on RoCoBench[155]. Qian et al. carefully divided the development process into four different chronological stages: design, coding, testing and recording. Each stage involves a team of agents, such as programmers, code reviewers, and test engineers, facilitating collaborative conversations and facilitating a seamless workflow. The chat chain acts as a facilitator, breaking down each stage into atomic subtasks[156]. Using multi-agent competition, the simple goal of hide-and-seek, and large-scale standard reinforcement learning algorithms, Bowen Baker et al. found that the agent created a self-supervised automated course, triggering multiple rounds of different emergent strategies, many of which required complex tool use and coordination[157]. Zhang et al. proposed a framework for multi-agent cooperation using LLMs. Experiments have proven that LLM-based agents that communicate verbally can win more trust and cooperate more effectively with humans[154].

- Self-Correction and Self-Improvement

The ability to automatically detect and analyze failed executions is critical for explainable and robust robotic systems. This requires the agent to self-criticize and continuously learn from previous tasks. Correct behavior within the bounds of common sense. Based on the large language model, directional error correction and directional improvement can be done. In task planning, it is necessary not only to have the current state, but also to have memory, experience, reflection and summary, and world knowledge. Agents can get feedback based on actions. For large language models, reinforcement learning from human feedback is an extremely simple environment.

Pan et al. analyzed and categorized a series of work of LLM's self-correction, including training time, generation time and post-correction[158]. Sharma et al. show how language can be used to update the latent cost of the planner to improve task performance. This method can use the language to correct the plan in two ways: adding constraints or specifying intermediate subgoals for the planner[159]. Liu et al. introduced a framework, REFLECT, to leverage the power of LLMs to explain robot failures. According to the explanation, the mission planner will generate an executable plan for the robot to correct the fault and complete the mission. Evaluating the framework on the RoBoFail dataset of failure scenarios, experiments demonstrate that the LLM-based framework is capable of generating informative failure explanations to aid in successful corrective planning[160]. Shinn et al. proposed a novel framework, reflexion, to learn from trial and error through dynamic memory and self-reflection to make better decisions in subsequent experiments[161]. Bousmalis et al. proposed RoboCat, a foundational agent for robotic manipulation, as a visual goal-conditional decision transformer. The authors demonstrate the ability to generalize to new tasks and robots. It also demonstrates the use of the trained model to generate data for subsequent training iterations, providing the basic building blocks for autonomous improvement loops[162]. Ning et al. explore whether LLMs has the ability to identify its own mistakes without resorting to external resources. In particular, the research focuses on whether they can be used to identify individual errors in step-by-step reasoning. The authors propose a zero-shot verification scheme to identify such errors. This verification scheme is then used to perform weighted voting on different generated answers, improving question answering performance[163]. Olausson et al. analyzed the ability of GPT3.5 and GPT4 to perform self-repair on APPS. The approach is to self- repair with separate code and feedback models. APPS is a dataset consisting of various coding challenges. By evaluating the strategy, it was found that the effectiveness of self- repair was only seen in GPT4. It was also observed that self- repair is bottlenecked by the feedback phase[164]. Continual learning serves as a means to consistently foster self-improvement and self-optimization[165]. Guo et al. proposed DoReMi, a novel language model base framework that can instantly detect and recover from inconsistencies between planning and execution[166]. In order to enable LLMs to integrate into the environment autonomously, Peng et al. proposed the self-driven grounding framework to automatically and gradually lay the foundation for LLMs through self-driven skill learning[167].

There are some other works that can be referenced[168-175], such as WebShop[176], InterCode[177], Collie[178].

### 3.3.3 Datasets and Metrics

**Datasets.** Researchers in the field of robot task planning rely on datasets to develop, validate, and test their planning algorithms and systems. These datasets may include real-world data collected from robots or simulated data generated within robot environments[179]. Here is an example list in Appendix Table 6.

**Metrics**. Task planning performance is evaluated by: (1) Task Success Rate. (2) Task Durations. (3) Task Diversity. (4) Task Difficulty. (5) Task Dependencies.

### 3.3.4 Problems

More effective strategies for self-improvement and enhancement need to be explored[180]. LLMs automatically detects when and how to apply the planner; LLMs reduces reliance on human information during planning[142]. In addition, the planning time efficiency of the model needs to be improved. The correctness and executability of generated plans need to be significantly improved. Multimodal models are explored for mission planning, which can naturally support extending the planning system to higher-dimensional observation spaces.

A significant advantage of utilizing the latest LLMs is their ability to adapt to various operating environments through several learnings and user feedback. These capabilities not only eliminate the need for extensive data collection or model retraining, but also allow users to make adjustments that promote safe and robust mission planning. The ability to effectively adapt to user feedback may be due in part to learning methods that combine model behavior with human intent. Additionally, the output of large models can be adjusted with a reasonable amount of feedback. The ability of LLMs to reflect the semantic content of user feedback provides means for users to communicate their intentions to the system. Therefore, this aspect helps lay the foundation for a user-friendly system. Delving into this ability to adjust contributes to the user-friendliness of the system[145].

## 3.4 Reasoning

### 3.4.1 Characteristic and Challenges

Reasoning requires the reasoner (robot) to have an explicit representation of various parts or aspects of its environment in order to reason. Robots are increasingly transitioning from specialized single-task machines to general-purpose systems operating in diverse and dynamic environments. To solve the challenges associated with real-world interactions, robots must effectively generalize knowledge, learn, and remain transparent in their decision-making processes. However, there are few studies that specifically address the reasoning challenges faced by robots in interactive environments. This survey aims to investigate reasoning robot system technologies that enable robots to encode and use knowledge, including concepts, facts, ideas, and beliefs about the world. Continuously sensing, understanding, and generalizing knowledge enables robots to identify meaningful patterns shared across problems and environments to perform a variety of real-world tasks more efficiently[19]. LLMs provide promising tools for robots to perform complex reasoning tasks[181]. In this section, we survey the recent research results and latest progress of foundation models in the direction of robot reasoning, including robot logical reasoning, common sense reasoning, affordance reasoning, and personalized reasoning[182-185].

### 3.4.2 Methods

- Logical Reasoning

Machine logical reasoning refers to the ability of a machine to draw new conclusions through the derivation of known facts and reasoning rules. In logical reasoning, the machine needs to take into account the logical relationship between facts, such as If A is true, then B is also true, One of A and B must be chosen, etc. In addition, machines need to be able to understand fuzzy information and take into account the possibilities of different situations and choose the most reasonable conclusion from them[186]. Chain of thoughts can significantly improve the complex reasoning capabilities of large language models[187]. Tree of thoughts can be used for tasks that require exploration, strategic foresight, or where initial decision-making play a key role[188].

Driess et al. implemented a PaLM-E multimodal model that plugs real-world continuous sensor modalities into a LLM to establish a link between words and perception. The inputs to the specific LLM are multimodal sentences that interweave visual, continuous state estimation, and textual input encodings. These encodings are trained end-to-end in conjunction with a pre-trained LLM to perform a variety of specific tasks, including sequential robot operation planning, visual questioning and answering, and captioning[189]. Brohan et al. argue that one of the keys to the success of general robotics models lies in combining high-performance architectures that learn a large variety of robotics data for training on open tasks[62]. Liang et al. proposed a method to generate policies based on LLMs. By inputting natural language instructions, the LLM trained by code completion can write robot policy codes. These policies are reactive policies and waypoint-based policies. Features include exhibiting spatial-geometric reasoning, generalizing new instructions, and assigning exact values to ambiguous descriptions based on context[168]. Yao et al. explore that with the support of chain of thoughts prompting and external knowledge bases, LLMs generates reasoning trajectories and task-specific actions in an interactive manner, thereby exerting greater synergy between the two[190].

There are still some valuable methods for reasoning that are worth studying, including accumulative reasoning[191], context learning[192] and causal reinforcement learning[193].

- Commonsense Reasoning

LLMs combined with knowledge graphs or langchain is a way to try common sense reasoning. LLMs are black-box models that typically fail to acquire and capture factual knowledge. In contrast, knowledge graphs etc. are structured knowledge models that explicitly store rich factual knowledge. Knowledge graphs can enhance LLMs by providing external knowledge for reasoning and interpretability. At the same time, knowledge graphs are inherently difficult to construct and evolve, posing challenges to existing methods for generating new knowledge and representing unseen knowledge in knowledge graphs. Therefore, it is complementary to combine LLMs and knowledge graphs to play their respective advantages at the same time[194]. LLMs and knowledge graphs (KGs) can complement each other, such that LLMs can be used for KG construction or completion, while existing KGs can be used for different tasks, such as making LLMs outputs interpretable or fact-checking in a neural-symbolic manner. Text2KGBench is a benchmark that evaluates the ability of language models to generate knowledge graphs from ontology-guided natural language text[195]. Kim et al. proposed KG-GPT, a general framework that utilizes LLM for knowledge graph reasoning[196]. Zhu et al proposed AutoKG, a method based on multi-agents, using LLM for knowledge graph construction and reasoning[197]. Yang et al. proposed to enhance the large language model KGLLM by developing knowledge graphs, and provided a solution to enhance the factual reasoning ability of LLMs[198]. Ren et al. propose KNOWNO, a framework for measuring and adjusting uncertainty in LLM planners so that they know when they don't know and seek help when needed. KNOWNO builds on conformal prediction theory to provide statistical guarantees for task completion. At the same time human assistance in complex multi-step planning setups is minimized[199]. Zellers et al. proposed QLeT: a model that learns physical commonsense knowledge through interaction and then uses this knowledge to construct language. Authors decompose QLeT into a physical dynamics model and a separate language model. Using a dynamics model as an interface to a language model, QLeT can read a sentence, neurally simulate what might happen next, and then communicate that result through a text-symbolic representation or natural language[200]. Zhu et al. proposed 3D-VisTA, a LLM with 3D world recognition capability, capable of answering questions based on a 3D world model. The project team also released the ScanScribe dataset, a 3D model-text dataset[201]. One approach to address inference problems is by incorporating vector databases. Both LlamaIndex and Langchain are working diligently to develop a data-augmented retrieval system, which could be further enhanced with a contextual agent. Yohei introduces the concept of incorporating relevant contextual information (task context), which may have nuances differing from those of the conventional semantic similarity algorithm offered by vector databases. Tang et al. proposed GraphGPT, which utilizes LLMs for graph instruction tuning[202].

- Affordance Reasoning

The ability to reason about affordances enables robots to choose actions that are appropriate for a given object and produce a desired effect. The Dreamer algorithm has recently shown great promise for learning from a small number of interactions by planning in a learned world model, outperforming pure reinforcement learning in video games. Learning world models to predict the outcomes of potential actions can be planned in imagination, reducing the amount of trial and error required in real settings[203]. Robots need primary knowledge of the world in which to act. LLMs can be used to score potential subsequent actions during task planning, or even directly generate action sequences without natural language instructions from additional domain information. Singh et al. propose a procedural LLM hint structure that enables plan generation to work across environments, robot functions, and tasks. LLMs are prompted by program-like specifications of operations and objects available in the environment, as well as example programs that can be executed. Situational awareness is introduced in LLM-based robot task planning[204]. Ahn et al. proposed SayCan, the value function of pre-trained skills to obtain the results of interaction with the environment, based on the real world. SayCan incorporates real-world experience into LLMs through the value function of pre-trained skills, enabling LLMs to execute real-world abstract, long-term commands on robots. This method implements language models to provide high-level semantic knowledge and provide pre-trained low-level skills to constrain the model to propose natural language actions that are both feasible and appropriate to the context[205]. Yoneda et al. published the Statler framework, which enables LLMs to have representations of world states that change over time while maintaining memory. The framework has two generic LLM instances: a world model reader and a world model writer. These two parts interact with and maintain the world state. With access to this world-state memory, the Statler framework improves the ability of existing LLMs to reason over longer periods of time, independent of context length constraints. Experiments on simulated domains and real robots show that the proposed method improves the state-of-the-art in LLM-based robot inference[206]. Gao et al. fine-tuned VLM on PhysObjects to improve its understanding of physical object concepts by capturing human priors on these concepts from visual appearance. Combining this physics-based VLM with a large language model-based robot planner into an interactive framework improves mission success rates[207]. Tang et al. proposed a knowledge condition detection framework CoTDet for affordance knowledge prompts for task-driven object detection[208]. Strategic robotic pursuit-avoidance requires exploiting the dynamics of interactions and planning through uncertainty in physical states and underlying intentions[209].

- Personalized Reasoning

The emergence of large language models marks a revolutionary breakthrough in artificial intelligence. A major leap forward in the capabilities of general artificial intelligence will change how personalization is implemented. On one hand, it will change the way humans interact with personalized systems. On the other hand, it will also greatly expand the scope of personalization[210]. The growing use of LLMs in conversational agents has sparked interest in the personality exhibited by data-trained models, as personality significantly influences communication effectiveness. Therefore, Safdari et al. proposed a comprehensive approach to test the personality traits expressed in the text generated by LLMs. Experiments find the reliability and validity of LLMs for simulating personality for larger and fine-tuned models. The personalities in the LLM output can be shaped in desired dimensions to mimic specific personality traits[211]. Huang et al. proposed to assess the empathic ability of LLMs, that is, how their feelings change when they encounter a specific situation. After experimental evaluation, LLMs can usually respond appropriately to some situations, although there are some biases. Still, they don't match human emotional behavior to make connections between similar situations. This paper expects to contribute to the advancement of LLMs to better adapt to human emotional behavior, thereby enhancing the practicality and applicability of intelligent assistants[212]. Wu et al. used TidyBot, a robot to learn personal preferences to personalize the cleaning of a room, where the robot used the planning and perception of language combined with the summarization capabilities of a LLM to pick up objects, determine where to place them, and organize the room[64]. Ding et al. Learn universal human priors for dexterous manipulation from human preferences[213]. Deng et al. present Socratis, a social response benchmark that tests the ability of state-of-the-art multimodal large language models to generate emotional reasons for a given IC pair[214]. Lian et al. proposed the first multimodal LLM in affective computing, called AffectGPT. The goal is to address the long-standing challenge of label ambiguity and chart a path toward more reliable technology[215].

### 3.4.3 Datasets and Metrics

**Datasets.** Robot reasoning datasets typically involve tasks that require problem-solving abilities such as logical reasoning and common-sense reasoning. These datasets are designed to evaluate the robot's ability to make inferences in various scenarios. Here are some examples of robot reasoning datasets in Appendix Table 7.

**Metrics**. The evaluation indicators for robot reasoning datasets typically depend on the specific dataset and task. However, common evaluation metrics and indicators for many reasoning datasets.

Reasoning performance is evaluated by: (1) Accuracy. (2) F1 Score: F1 score is calculated based on the overlap between predicted answers and ground truth answers. (3) Exact Match (EM): Calculates the percentage of questions answered exactly correctly. (4) Top-k Accuracy. (5) Mean Reciprocal Rank (MRR): Measures the quality of the top-ranked answer. It calculates the average reciprocal rank of the first correct answer in the ranked list of answers. (6) BLEU (Bilingual Evaluation Understudy): Commonly used in machine translation tasks, BLEU measures the similarity between predicted and reference answers.

### 3.4.4 Problems

Robot learning goals are to master how to learn, combine advanced pattern recognition with model-based reasoning, and develop common sense intelligence. With the advancement of learning and the improvement of intelligence[216], research on robot reasoning has gradually deepened. Reasoning is an abstract, advanced form of thinking. The objective basis of reasoning is the relationship between objective things. At present, the reasoning capabilities of robots are weak (including causal reasoning[217], spatiotemporal reasoning, real-time reasoning, geometric reasoning, world model, world knowledge, common sense reasoning, understanding of physical constraints, etc.). One very important point is that the robot lacks reverse reasoning, and the basic model may reason from left to right. The current model is based on text-visual speech, so it does not directly reason about information such as touch, and more powerful multi-modal models are needed when available.

Solutions that can be tried are as follows: Beyond statistical correlation, reasoning about system dynamics and causality; meta-learning with limited data; rapid learning to adapt to dynamic, uncertain environments; learning across heterogeneous tasks and domains ; Developing systems that know their limitations and know how to ask for help; Developing systems that can deeply understand and synthesize complex textual and narrative information; Conducting deep moral and social reasoning about real-world problems.

## 4 Discussion

Despite the significant progress made in robot learning based on LLMs techniques, the field still faces numerous challenges in terms of both technical aspects and ethic aspects. In the following section, we will outline the major challenges, potential solutions as well as potential future directions. We hope that the highlight aspects can serve as inspiration for future research investigations in the robot learning area.

### 4.1 Robot Hardware and Software Decoupling

The basic requirement for a foundation model is a unified architectural framework. However, there is significant variation in robot hardware, making it crucial to achieve bottom-level uniformity or decoupling. To facilitate the advancement of robot technology, it is crucial to ensure the synchronization of form with function[218], which implies that software and hardware must evolve simultaneously. Nevertheless, numerous challenges persist in the development of software and hardware. On one hand, due to the diversity of manufacturers' specifications and private parameters, it becomes challenging to individually program and manage robots when (re)configuring them to achieve desired tasks or formations. This dilemma often occurs in industrial and operational fields. On the other hand, the construction of current robot models and databases largely depends on the hardware structure of the robots. The majority of robot databases are constructed by collecting specific data pertaining to individual robots. Consequently, trained robot models exhibit optimal performance solely on the particular robot they are trained for. Put simply, existing robot models are limited by the hardware structure of the robots and software algorithms.

Ideally, there is a need to decouple robot software and hardware, and concurrently, collect databases encompassing various robot models. The separation of the logical and physical components would foster increased software innovations and facilitate a potentially more accessible robot market, inevitably resulting in reduced non-recurring engineering costs. In a decoupled architecture, robot hardware and software can be developed and updated independently, without restrictions. Additionally, the most recent high-performance software algorithms can be efficiently implemented across various robot models. Confronted with diverse hardware configurations and unpredictable operating environments, users can develop and program robots of different types without the need for a comprehensive understanding of the specific hardware employed by each robot. Compatible APIs, predefined libraries, and programming software can be developed and enhanced. The abstraction level of robot software needs to be improved to increase the efficiency and effectiveness of robot operations, enabling robot applications to run robustly in dynamic environments[219].

## 4.2 Dynamic Data for Interaction with the Environment

When establishing a specialized large-scale robot model, it is necessary to utilize dynamic and diverse data[220], including dynamics data, during the training or fine-tuning process. Meanwhile, robots are anticipated to embody key attributes including agility, cost-effectiveness, diversity, environmental adaptability, and plasticity, which empower them to execute tasks like fastening garments and tying shoelaces. Interactive environments[221] and abundant dynamic data are imperative for sufficiently training and assessing robots possessing these aforementioned attributes. However, the realization of authentic robot scenarios and the acquisition of real-time data present formidable challenges. Therefore, collaboration and resource sharing among laboratories worldwide assume paramount importance in propelling responsible and open advancements in robotics research. Data collection methods may encompass several approaches, including action library aggregation, teleoperation, and imitation learning. The crux of the challenge lies in dissociating the collected data from the specific robot model employed, thus ensuring its perpetual validity. Additionally, the existing training process in robot learning suffers from an inadequacy of effective and abundant data. Establishing multimodal databases that can be utilized for robot learning training is of paramount importance.

In order to better understand the environment, recognize objects, and perform tasks, current robot models often only focus on the fusion of top-level data such as vision and language in multimodal fusion, while neglecting bottom-level data such as dynamics data related to interaction with the environment. In real-world environments, there is no doubt that mechanics directly determine the stability of grasping. To improve the accuracy, generalization, robustness, and address issues related to force perception in robot models, it is necessary to construct large-scale dynamics data during training. Dynamics data includes information such as position (XYZ, rotation angles), acceleration, and forces (magnitude and torque), etc. Based on this information, robots can easily manipulate objects of different materials and weights accurately. Understanding the robot's dynamics model allows the model to comprehend the robot from a bottom-up approach during the training process of existing robot learning. Incorporating dynamics data into the robot's foundation model also enables the model to generate more realistic and physically meaningful data.

## 4.3 Robot Generalization

Robot models need to demonstrate better generalization ability, surpassing the semantic, visual, and other multimodal understanding of the robot data they encounter. It requires robots to perform operational tasks on objects or scenes in the robot data that they have never seen before. This necessitates the utilization of knowledge derived from network data for operation. Robot models should not only maintain performance on the original tasks in robot data but also improve performance in previously unseen scenarios. Accordingly, we present the following key elements.

Currently, the advancements in robotic research have yet to reach optimal levels of efficacy. A clear divergence is evident between human intelligence and that of robots, especially when evaluating areas such as image recognition and the competency of platforms like ChatGPT in question-answering scenarios. The enhancement of both the precision and efficiency of robotic learning stands as a critical imperative. While modern robotic learning methodologies have attained certain milestones, the scope for refining execution success rates remains substantial.

The environment is expected to become increasingly complex. Robots need to adapt to various scenarios to perform a variety of tasks. It is difficult to achieve long-term autonomy in complex environments, so lifelong learning is required. Currently, robots are only capable of performing a limited number of skilled tasks and are not generalizable across a variety of tasks[222]. On the one hand, Robots will adapt to different scenarios including layout configuration, visual texture, light source changes, time simulation, low cost, data enhancement, demand personalization, and program automation. In virtual-real transfer, methods that may be used include domain adaptation, meta-learning, transfer learning, knowledge distillation, and world models. Legged robots have made a lot of progress in traditional methods, but they are lacking in combining with foundation models. On the other hand, Robots can observe, understand, practice in interaction with their environment, and thus self-exam, diagnose, and repair. The robot can not only collect data by itself, analyze the data, and then analyze the cause of the failure and then solve the problem. Robots can have self-driven learning awareness and can continue to learn[165].

Currently, foundation models are developed and trained in unmanned environments. In the future, when they are utilized in environments with human presence, interactions between robots and humans will need to take human factors into consideration. It requires human-robot alignment and considers emotional factors. It also considers biomimetic learning, cross-learning, and the co-development of the brain and the body.

AI alignment is worthy of in-depth research. The goals of AI systems are required to be consistent with human values and interests. If the values of AI and humans cannot be aligned, the following problems will arise. Behaviors that do not meet the goals of human intentions, out of control, harming human interests, and making wrong choices in the conflict of multiple setting goals[157, 223].

Researchers can combine functionality with expressive capabilities and explore methods to improve robots from this perspective. Functionality is manifested in mobility, dexterity, perception, and intelligence. Considering the relationship between individuals and occupations, we can liberate people from dangerous situations and enhance the technological level of humanity.

While current foundation models are capable of engaging in high-level semantic conversations, they fall short of humans in low-level control aspects such as movement and operation. Consequently, bionics is deemed significant. Biomimetic learning is a highly worthwhile field of research in order to design and develop more powerful robots. It draws inspiration from biological systems, especially animals and organisms. By observing how animals and organisms move, perceive their environment, and interact with it, researchers attempt to replicate these mechanisms in robots. They utilize biomimetic materials, such as artificial muscles and soft robot components, to enable robots to imitate the adaptability and flexibility of living organisms. The objective of researchers is not only to replicate the physical characteristics of animals but also to understand their behavioral patterns. This includes understanding how animals make decisions, solve problems, and exhibit intelligent behavior. Overall, the aim of biomimetic learning in robotics is to push the limits of robot capabilities by harnessing natural wisdom and integrating biological principles into the design and control of robots, making them more versatile and adaptable to various tasks and environments.

Robots can achieve dexterous movement and manipulation. It has been found through research that the perception-action circuit is the center of cognition, and the body uses the perception-motor system to generate intelligence in the interaction with the environment. Intelligent robots span multiple directions in intelligent disciplines, such as cognitive science, psychology, brain science, and sociology. Therefore, the focus is on interdisciplinary integration, analogous to the developmental process of human intelligence. In multi-modal environmental interaction, by opening up the links and loops between morphology, perception, behavior and learning, robots can realize active perception and autonomous learning.

Being conscious is hard, but the body and brain of robot should develop together[218, 224]. In robot learning, the robot lacks common sense and lacks the physical constraints of learning real scenes from the environment. Robot learning is to make robots intelligent. In addition to perceptual and motor capabilities, which have been studied in the past, cognitive abilities are also very significant. Referring to the development process of human intelligence, the brain and body develop simultaneously. Intelligent robots not only have perception and motor intelligence, but also cognitive intelligence with autonomous consciousness. In addition, robots lack the ability to learn actively. For example, in completing the specified target tasks, it can actively ask what it needs to do next in order to achieve a certain task. Active learning may regain favor as autonomous contextual agents actively reveal what they don't know in order to prioritize[225].

## 4.4 Multimodal Interaction

Currently, the interaction data of robots mainly relies on visual and textual information, lacking diversity. In order to enhance the robot's perception of the environment and its interaction with it, it is necessary to integrate multimodal data from various sensory modalities, including visual, auditory, tactile, olfactory, gustatory, and other sensory inputs. Although multimodal data brings great hopes for advancing robot technology, it also comes with some problems and challenges. Specifically, different modalities have different data formats, structures, and features, posing challenges for the collaborative interaction among different modalities. The main issues involved are multimodal representation, multimodal mapping, multimodal alignment, and multimodal fusion.

To address these problems, several considerations can be taken into account. Research can be conducted to tackle challenges related to large-scale environments, multitasking, and strong interaction. Tasks with higher interactivity can be verified and demonstrated in more open and complex environments, such as simulators and multi-platform settings. Various fusion strategies can be explored to effectively combine information from different modalities, including hierarchical and multilingual fusion, as well as scalable fusion. Advanced deep learning models, such as multimodal transformers, can be developed to efficiently fuse and extract meaningful information from different sensory modalities. Cross-modal learning techniques can be explored to enable robots to learn meaningful correlations between different modalities, thereby improving their understanding of the environment. Transparent and interpretable multimodal fusion models can be developed to facilitate human understanding and trust in robot decisions. The capability of robots to understand the semantic of multimodal data, including scene understanding[15] and context comprehension, can also be enhanced. Additionally, during the process of converting multimodal content, foundation models may lose information or generate errors, leading to biased results. This requires foundation models to achieve strong multimodal understanding, addressing the current issues of unreliability in intelligent agents, as well as information loss and biases.

## 4.5 Exclusive Foundation Models for Robots

Foundation models have achieved many gratifying results in text, language, etc., but in the field of robotics, the development of foundation models is slow. There is an urgent need to develop usable and effective general-purpose foundation models and special-purpose foundation models of robots. In order for robots to reach human-like levels of capability, it is necessary to collect robot data for every object, environment, task, and situation. In addition to the issue of robot data, the following problems and challenges also need to be overcome.

Multimodal large language models suffer from the catastrophic forgetting problem. Fine-tuning of a model can improve specific performance, but as fine-tuning proceeds, the model begins to exhibit hallucinations, resulting in a significant loss of generality [226]. The hallucination problem of foundation models, that is, when the text generated by the model does not follow the original text (Faithfulness) or does not conform to the facts (Factualness)[227]. As a result, the robot gives wrong judgments or operations in the process of performing tasks in combination with the foundation model[228]. Currently, in order to obtain reasonable answers in multi-turn dialogues with foundation models, multiple prompts need to be provided. Therefore, it is necessary to design foundation models that do not rely on human prompts. Furthermore, foundation models need to have better contextual understanding and causal understanding[229, 230].

Combining large models and small models is also a method that can be tried [231]. Large models are well suited to handle Corner Cases, while small models trained by modeling and for scenarios cannot exhaustively cover the whole scenario, and this part of the work can be handled by the general understanding and strong generalization and inference capabilities of large models. The combination of large model and small model can improve the inference efficiency without reducing the inference ability.

Foundation models for low level control need to be designed. Due to the lack of data for low-level controllers in the training corpora, most of the existing robot foundation models have been applied to robots as either a semantic planner or to interact with robots using human-designed action primitives[232]. Having a large embodied model enables one-step control of the lowest level.

The evaluation of embodied intelligent agents should shift from task-oriented evaluation to capability-value evaluation. Human discrimination tests evaluate AI based on human observations, as represented by the classic Turing test.

Qualitative testing can only be done through human observation and there is no quantitative testing[233-239].

4.6 Computation Efficiency

Computing efficiency[240] requires continuous evolution and improvement. Optimization aspects while ensuring high performance mainly include the following three aspects: computing costs need to be reduced, computing resources need to be reduced, and computing time needs to be reduced.

There has been a staggering increase in computational power requirements. However, storage performance significantly falls behind that of processors. These two factors contribute to the problems of the computational wall and the storage wall. In response to the above challenges, researchers are committed to addressing issues related to novel AI storage and computing technologies. These technologies aim to break through the storage and computational bottlenecks of AI calculations, improve computational efficiency, and mainly include new applications, computing frameworks, storage-computing architectures, and cloud infrastructure technologies. In terms of hardware architecture, there are two aspects: von Neumann architecture and novel storage-computing architecture. The novel storage-computing architecture can focus on integrated storage-computing chips, neuromorphic chips, and so on.

The models involved in the robot system require efficient computing, including the following aspects: cloud-edge-device integration and smart chips. Cloud-edge-device integration requires unified management, cloud-edge collaboration, and resource allocation. Smart chips need to be small in size, low in power consumption, and high in performance.

The existing foundation models have a huge amount of parameters, and the training of parameters requires the support of many high-performance graphics cards, which requires a huge cost. In order to reduce the cost and the amount of parameters of the model, a lightweight foundation model is designed while ensuring the performance[241]. Additionally, further compression can be performed on the network structure of large-scale models. In addition to adjusting the parameter size, extraction methods such as knowledge distillation, network pruning, low-rank parameter decomposition, and quantization can also be employed.

4.7 Robot Safety and Ethics

Robot security includes two aspects: physical security and data security. Physical safety refers to the fact that some misunderstandings when combining LLMs with robots may lead to unexpected chain reactions. For example, the robot received instructions to cook Western food in the oven, but it cooked Chinese food. As a result, it turned on the gas and accidentally caught fire. The specificity of robot tasks lies in the constant interaction with the environment during task execution, and robot safety becomes particularly crucial when humans are present in the environment. In the process of completing the specified target tasks, the robot lacks safety guarantee due to physical constraints and other real environment restrictions, and there may be accidents such as collisions, extrusions, and damage to mechanical parts, potentially causing harm to humans.

There is also data security, which involves data privacy. Privacy risks in robot foundation model development and application primarily come from the information contained in the original training data and the powerful inference capabilities of the models. Developers need to ensure that robot foundation models do not cause privacy breaches and carefully evaluate the potential ethical issues they may bring about. In addition to the data bias in the LLMs being trained, every user will also have information security concerns when training the foundation model and uploading the data to the cloud. The intellectual property rights and data security cannot be guaranteed. This requires the use of data desensitization. In addition to the risks of sensitive information leakage and content infringement, the following risks also exist: model denial of service, model theft, training data poisoning, model hallucination, and attacker prompt injection. Therefore, it is necessary to establish a comprehensive and multi-faceted security evaluation, detection, and defense system.

Ethically, the behavior of robots should comply with social and legal norms. Developing autonomous robots with self-awareness is also a promising direction. There is growing concern about ethical and safety aspects of robotic learning, including fairness, transparency, and robustness. It is important to pay attention to biases and toxicity in the training samples of robot foundation models to ensure that robots' behavior does not lead to discrimination or unfairness.

## 5 Conclusion

This paper provides an overview of the key challenges of robot learning and the types of algorithms that combine robot learning with foundation models developed to address these challenges. We outline the development and evolution of related technologies for robot learning, as well as the prerequisites such as datasets and computing resources required. We divide these key robot learning challenges into four categories according to downstream tasks, namely manipulation, navigation, planning, and reasoning. With the development of foundational models, they have demonstrated significant progress in robot applications and promising humanoid intelligence[242]. These findings present a bright future for foundational models in robot applications. Last but not least, discussions were conducted, which explained the current problems and challenges of robot learning, and proposed research directions in the future, including robot hardware and software decoupling, dynamic data for interaction with the environment, exclusive foundation models for robots, and so on.

## Appendix

Table 1. Simulation Framework

| | MineDojo[243] | Habitat 2.0[244] | Habitat3.0[245] | BEHAVIOR-100[246] | BEHAVIOR-1K[247] | iGibson 1.0[248] | AI2-THOR 2.0[249] | BabyAI[250] | PyBullet | PyRobot[251] | Isaac Sim[252] | RFUniverse[253] | Unisim[254] |
|---|---|---|---|---|---|---|---|---|---|---|---|---|---|
| Simulator | MineDojo | Habitat Sim | Habitat Sim | iGibson 2.0 | OMNIGIBSON | iGibson | AI2-THOR 2.0 | MiniGrid | PyBullet | Gazebo | Omniverse | RFUniverse | Unisim |
| Dataset | MineDojo | ReplicaCAD | Habitat Synthetic Scenes | Human VR demos | BEHAVIOR-1K | iGibson dataset | iTHOR, RoboTHOR, Proc | | | | | | |

| | | | | | | | | | | | | | | |
|---|---|---|---|---|---|---|---|---|---|---|---|---|---|---|
| | | | Dataset | | | | THOR-10K, ArchitecTHOR | | | | | | | |
| Sensors/Sensor Signals | Video | RGB-D Cameras, Joint-position sensors, Egomotion sensors | RGB-D Cameras, GPS | On-board virtual sensors (RGB, Depth images, LiDAR, Normals, Flow (Optical, Spatial), and Semantic and instance segmentation) | On-board virtual sensors (RGB, Depth images, LiDAR) | RGB images, Rendering of normals, Depth, Point clouds, Virtual LiDAR signals, and Optical scene flow | Cameras and the environment meta data | | | Camera, Distance, Proximity sensors, Laser, force sensors | RGB-D, Lidar, and IMU | Vision, IR, DIGT | Cameras, LiDAR |
| Language | JAVA, Python | C++, Python | C++, Python | Python | Python | Python, C | C#, Python | Python | Python | C++, Python | Python | C# | |
| Supported OS | Linux/Mac OS/Windows | Linux/Mac OS/Windows | Linux/Mac OS/Windows | Windows/Ubuntu | Windows/Ubuntu | Windows/Linux/Mac | Mac OS, Ubuntu | Windows/Linux/Mac | Windows/Linux/Mac OS | GNU/Linux (Ubuntu) | Windows/Linux | Windows/Linux | |
| Supported Task | Minecraft Game | Navigation, Manipulation | Navigation, Manipulation | Navigation, Manipulation | Household Activities | Navigation, Manipulation | Navigation, Manipulation | Navigation, Manipulation | Manipulation, Locomotion, Control | Navigation, Manipulation | Navigation, Manipulation | Complex Dynamics Tasks | Self-Driving |
| Open Source | Open | Open | Open | Open | Open | Open | Open | Open | Open | Open | No | Open | No |
| Backend | Pytorch | | | | | | | Pytorch | | | | | |
| Scenes | | 105 | 211 | 15 | 50 | 15 | 120 | 19 | 9 | | | | 103 |
| Objects | | 92 | 18k | 391 | 5000+ | 570 | 84 | | | | | | |
| Agent/World Interaction | | | | Mass, Center of Mass, | | Force | | | | | | | |
| | | | | Friction | | | | | | | | | |
| Environment | 3D | 3D | 3D | 3D | 3D | 3D | 3D | 2D Gridworld | | 3D | 3D | 3D | 3D |
| Physics Engine | Minecraft game | Bullet | Bullet | PyBullet | PhysX 5 | PyBullet | Unity | | Bullet Physics SDK | ODE, Bullet, Simbody, DART | PhysX | Unity | |
| 3D Rendering Engine | Minecraft game | Magnum | Magnum | Physics-Based Rendering | Nvidia Omniverse | Physics-Based Rendering | Unity | | OpenGL | OGRE | RTX | Unity | |
| Speed | | 1400 steps/s | 140-250 steps/s | | | 100 steps/s | 90-180 steps/s | | Each time step in PyBullet is 1/240 seconds | | | | |
| Robot Family | | Articulated Robots | Spot Robot | A Bimanual Humanoid Avatar, A Fetch Robot | | Articulated Robots, Fetch Robot | | | R2d2 Robot | Mobile, Humanoid, Industrial | A Wheeled Robot and A Frank, a Emik, a Robotic Arm | | |
| Supported Tools | | Grippers, Arm Manipulator | | | | Grippers, Locobot | | | | Grippers | Manipulators | | |
| Computing Resource | 8× V100 GPUs | 8GPU | | | | | | 20~50GPU | A high-end desktop GPU | | | | |

Table 2. Comprehensive Robot Datasets

| | Data Type | Objects | Scenes | Data Volume | Task | Skills |
|---|---|---|---|---|---|---|
| Open X-Embodiment[255] | Demonstrations | | | 4435.41GB | Manipulation | 527 |
| HoloAssist[256] | RGB, depth, head pose, 3D hand pose, eye gaze, audio, and IMU, text | | | 166h | Collaboratively manipulation | 20 |
| UniMoCap | Text-motion mocap | | | 34K motions and 66K annotations | Describing the actions being performed in mocap sequences | |
| ARMBench[257] | Images and videos | 190K | 1 | 235K | Object segmentation, Object Identification, and Defect Detection | |
| RT-1[62] | Instruction and image tokenization | | 3 | 130K | Manipulation | 744 |
| RoboTurk[258] | Demonstrations | | 2 | 137h | Block Lifting (lifting), Bin Picking (picking), and Nut-andpeg Assembly (assembly) | |
| Raven[259] | Images and RPM | | | 1120000 images and 70000 RPM | Reasoning, VQA | |
| RoboNet[260] | Video frames | | 4 | 15 million video frames | Manipulation | |
| GSO[261] | 3D images | 1030 | | 13G | Manipulation, Navigation and so on | |
| Meta-world[262] | Video, trajectory | | 50 | | Manipulation | 50 |
| RLBench[263] | RGB, depth, and segmentation masks | | | | Manipulation | 100 |
| M2DGR[264] | RGB image, 3D point cloud, inertial data, GNSS signals | | | 36 sequences (about 1TB) | SLAM | |
| OBJECTFOLDER 2.0[265] | Object files (containing the complete multisensory profile) | 1000 | | 1000 object files | Object scale estimation, Contact localization, and Shape reconstruction | |
| Google Brain Robot Data[266] | Images | | | ~800k grasp attempts | Manipulation | |
| Dex-Net 2.0[267] | Point cloud | 1500 | | 6.7 million point clouds | Manipulation | |
| Bridge Data[268] | Demonstrations (video) | | 10 | 7200 demonstrations | Household kitchen tasks | 71 |
| RH20T[269] | RGB image, Depth image, Binocular IR images, Robot joint angle, Robot joint torque, Gripper Cartesian pose, Gripper width, 6-DoF Force/Torque, Fingertip tactile | | | ~110K | Learning task and motion planning | 147 |
| Radish | Odometry, laser and sonar data, sensor data, Environment maps | | | Multi datasets (Usc Sal200 Synthetic, Robonaut Sensor) | Robotics dataset community | |
| Daily Interactive Manipulation[270] | Position, orientation, force, and torque of objects | | | 3354 trails | Manipulation | 33 |
| Robot @ Home[271] | Intensity images, depth images, and 3D point clouds | | | 9.6G | Object/room instance recognition, object segmentation, data | |
| | , laser scanner data, topological information | | | | compression/transmission | |
| TEACh[272] | Language | | | 3047 sessions | Navigation, Dialogue | |
| Robotic 3D Scan Repository | 3D point clouds | | | Multi datasets (FHG Campus, FireAcademy) | SLAM (navigation) | |
| MRPT | Sensors from 2D laser scanners up to RTK GPS, stereo cameras or 3D ToF cameras | | | Multi datasets (Kenmore, Edmonton 2002) | Mobile robotics and computer vision | |
| ImageNet[273] | Image | | | 1.4 million+ images | Computer vision | |
| EgoNet[273] | Image | | 100+ | 1.5 million video frames | Manipulation | |
| OakInk-Image[274] | Image | 100 | | 230K image frames | Hand Mesh Recovery and Hand-Object Pose Estimation | |
| OakInk-Shape[274] | Obj models | | | 62K hand-object poses and models | Grasp Generation, Intent-based Interaction Generation, and Handover Generation | |
| HANDAL[48] | Image frames | 210 | | 306K image frames | Manipulation | |
| ScanScribe[201] | 3D scan and text | 56.1k | 1,185 | 2995 RGB-D scans | 3D vision-language grounding | |
| Sound-Action-Vision Dataset[105] | Sound, RGBD, tracking location | 60 | | 15000 interactions | Interplay of action and sound | |

Table 3. Foundation Models

| | Foundational Model | Data Type | Data Size | Parameters Scale | Tokens Scale | Open Source | Training Time | GPU Numbers | Publisher |
|---|---|---|---|---|---|---|---|---|---|
| Language Models | ChatGLM[275] | Text | 1.2T English, 1.25T Chinese (130B) | 6B-130B | 400B-1T | ✓ | 60 days | 96 DGX-A100 GPU (8×40G) | Tsinghua |
| | T5[276] | Text | | 60M-11B | 1T | ✓ | | | Google |
| | GPT3[277] | Text | | 175B | 300B | ✗ | | | OpenAI |
| | LaMDA[278] | Text | 1.56T words | 2B-137B | 2.81T | ✗ | 57.7 days | 1024 TPU-v3 chips | Google |
| | LLaMA[279] | Text | 4.7T | 7B-65B | 1.4T | ✓ | 21 days | 2048 A100 GPU | Meta |
| | MOSS[280] | Text | 700B words | 16B | | ✓ | | | Fudan University |
| | InternLM[281] | Text | 1.6T | 7B, 20B, 104B | 1.6T | ✓ | | | Shanghai AI Laboratory, SenseTime |
| | Baichuan 2[282] | Text | 127M vocab size | 7B, 13B | 2.6T | ✓ | | 8 A800 GPUs | Baichuan Inc. |
| | Qwen[283] | Text | 3T | 7B, 14B | 3T | ✓ | | | Alibaba |
| | OpenChatKit | Text | 43M instructions | 20B | | ✓ | | | Eleuther AI |
| | PANGU-Σ[284] | Text | 2158 GB | 1.085T | 329B | ✗ | 100 days | 512 Ascend 910 NPUs | Huawei |
| | ColossalChat[285] | Text | 104K Chinese and English. | 7B | 54M | ✓ | | 8 A100-40G GPUs | UC Berkeley |
| | Alpaca[286] | Text | 52K instructions | 7B | | ✓ | 3 hours | 8 80GB A100s | Stanford University |
| Vision Models | SAM[287, 288] | Image | 11M images, 1.1B masks | 91M-636M | | ✓ | 68 hours | 256 GPUs | Meta |
| | DINOv2[289] | Image | 1.2B images | ~1B | | ✓ | 2 days | 8 V100-32GB GPUs | Meta |
| | ViT[290] | Image | 317M images | 86M | | ✓ | | | Google |
| | VideoMAE V2[291] | Video | 1.35M clips | 1B | | ✓ | 14 days | 64 A100 GPUs | Nanjing University |
| Multilingual Models | BLOOM[292] | Text | 341B tokens of text | 176B | 366B | ✓ | 3.5 months | 48 nodes with 8 NVIDIA A100 80GB GPUs | Google |
| | Vicuna[293] | Text | 70K samples | 13B | | ✓ | 1 day | 8 A100 GPUs | UC Berkeley, etc. |
| | Claude | | | | | ✗ | | | Anthropic |
| Multimodal Models | GPT-4[294] | Text, Image | | 175B | 13000B | ✗ | | | OpenAI |
| | GPT-4V[295] | Text, Image | | | | ✗ | | | OpenAI |
| | MiniGPT-5[296] | Text, Image | | 7B | | ✓ | | 4 A6000 GPUs | University of California, Santa Cruz |
| | ERNIE Bot[297] | Text, Image | 4TB Chinese text corpora | 100B-260B | | ✗ | | | Baidu |
| | CLIP[298] | Text, Image | 400M (image, text) pairs | ~1.6B | | ✓ | | | OpenAI |
| | DALL-E[299] | Text, Image | 250M text-images | 12B | | ✓ | | | OpenAI |
| | ViLT[300] | Text, Image | 4M images and 9.8M captions | 87.4M | | ✓ | | 64 NVIDIA V100 GPUs | NAVER AI Lab |
| | VisionLLM[301] | Text, Image | 238K images | 7B | | ✓ | | 4 × 8 NVIDIA A100 | Shanghai AI Laboratory |
| | Prismer[302] | Text, Image | 11M images, 12.7 M(images, text) | 275M-1.6B | | ✓ | | | Imperial College London, etc. |
| | BEiT-3[303] | Text, Image | 15M images and 21M image-text pairs | 1.9B | | ✓ | | | Microsoft |
| | LLaVA[304] | Text, Image | ~80K images | 13B | | ✓ | | | University of Wisconsin–Madison |
| | USM[305] | Text, Audio | 12M hours of speech, 28B sentences of text | 2B | | × | | | Google |
| | AudioPaLM[101] | Text, Audio | | 600M, 2B | | × | | | Google |
| | Whisper[306] | Text, Audio | 680,000 hours of audio | 39M-1550M | | ✓ | | | OpenAI |
| | MMS[307] | Text, Audio | 107.4 h | 0.3B, 1B | | ✓ | 2.3h(0.3B) or 3.5h(1B) | 48(0.3B) or 64(1B) A100 GPU | Meta |
| | DALL • E3[308] | Text, Image | 1B images | | | × | | | OpenAI |
| | Meta-transformer[309] | 12 Types (Text, Images, Point Clouds, Audio, Video, etc.) | Total ~ 449M | 85M, 302M | | ✓ | | | Chinese University of Hong Kong, etc. |
| Embodied Agents | Palm-E[189] | Text, Image | | 12B, 84B, 562B | | × | | | Google |
| | VIMA[310] | Text, Image, Video | 650K trajectories | 2M-200M | | ✓ | 1 days | 8 NVIDIA V100 GPUs | Stanford University |
| | RT-2[102] | Text, Image, Action | 10B image-text pairs | 5B, 55B | | × | | | Google DeepMind |
| | RT-X[255] | Text, Image, Action | | 35M, 5B, 55B | | ✓ | | | Google DeepMind |
| | LM-Nav[311] | Text, Image, Action | | (Based on GPT3, ViNG, CLIP) | | ✓ | | | UC Berkeley |
| | VoxPoser[312] | Text, Image | | (Based on GPT4) | | ✓ | | | Stanford University |
| | VOYAGER[313] | Text, Image | | (Based on GPT4) | | ✓ | | | NVIDIA |
| | DIAL[69] | Text, Image | 80k trajectories | (Based on CLIP) | | × | | | Google |
| | Gato[314] | Text, Image, etc. | 63M Episodes | 1.2B | 1.5T | × | | | DeepMind |
| | RoboCat[162] | Text, Image, etc. | | 1.2B | | ✓ | | | DeepMind |
| | MOMA-Force[99] | Vision, Force | | | | × | | | ByteDance |
| | Co-LLM-Agents[154] | Text | | (Based on GPT4) | | ✓ | | | University of Massachusetts Amherst |

| | | | | | | | | |
|---|---|---|---|---|---|---|---|---|
| | CLIPORT[71] | Text, Image | 200K iterations | (Based on CLIP) | | ✓ | 1 day | 6 GPU | UW&NVIDIA |
| | Instruct2Act[315] | Text, Image | | (Based on SAM and CLIP) | | ✓ | | | Shanghai Jiao Tong University |

Table 4. Manipulation Datasets

| | Objects | Scene | RGB-D Images | Depth Images | 3D point clouds | Size | Num of grasp | 6 DoF | Multi-cam | Real Image | Shape completion | Humanoid gripper |
|---|---|---|---|---|---|---|---|---|---|---|---|---|
| DailyGrasp[316] | 26 | | 26005 | 26005 | | 26K motion steps | | 7DoF | ✓ | ✓ | | ✗ |
| Standford Grasping[317] | 10 | | 13747 | 13747 | | | | ✓ | | | | |
| Cornell Grasping[318] | 240 | | 885 | 885 | | 7200 grasps | | ✓ | | ✓ | | ✗ |
| YCB Benchmarks[319] | 77 | | 46200 | 46200 | ✓ | | | ✓ | ✓ | ✓ | | ✗ |
| CMU dataset[320] | 150+ | | 50567 | | | About 1G | 50K | 7DoF | ✓ | ✓ | | ✗ |
| Google grasping dataset[266] | | | 800000 | | | 650K | | 7DoF | ✓ | ✓ | | ✗ |
| Dex-Net 4.0[321] | 25 | | 5M | 5M | ✓ | 103.16G | | | ✗ | ✗ | | ✗ |
| JACQUARD[322] | 11k | | 54485 | 108970 | | | 1.1M | | ✗ | ✗ | | ✗ |
| GraspNet-1Billion[323] | 88 | 190+ | 97280 | | | 2.1B | 1B+ | ✓ | ✓ | ✓ | | ✗ |
| kPAM-SC[324] | 33 | 125 | 111k+ | | ✓ | 111k | | 7DoF | ✗ | ✓ | ✓ | ✗ |
| Min Liu et al.[325] | 324 | | | | | | | 25DoF | ✗ | ✗ | | ✓ |
| HOI4D[326] | 800 | 610 | 2.4M | 2.4M | ✓ | | | | ✓ | ✓ | | ✓ |
| DexGraspNet[83] | 5355 | | | | | | 1.32M | | | ✗ | | ✓ |
| Grasping-anything[327] | ~3M | | | 1M | | 1M samples | ~600M | | | ✗ | | |

Table 5 Visual Language Navigation Datasets

| | Data type | Size | Scenes | Annotation | Objects | Instructions | Buildings |
|---|---|---|---|---|---|---|---|
| NavigationNet[328] | Image | Hundreds of thousands of images | 15 | | | | Each scene with 1-3 rooms |
| M2DGR[264] | Image, IMU, VI sensor, LiDAR, GNSS | 1220.6GB | 36 | | | | |
| DISCOMAN[329] | Image, depth, IMU, grid | 200 long sequences | | | | | |
| ALFRED[138] | Images, instructions | 8055 expert demonstration episodes | 120 | 25k | | ✓ | |
| VXN[330] | Image, text, audio | 9.6M episodes | 58 | | 21 (category) | | |
| SCAND[331] | Rosbags, LiDAR, Images, Visual Odometry, Wheel odometry | 8.7hour/138 trajectories | | 1656 | | | |
| MuSoHu[332] | Image, LiDAR, IMU, audio, depth | 300 trajectories | | 17 labels | | | |
| R2R[139] | Image, text | 7189 trajectories | 90 | | | 22k | |
| BnB[333] | Image, text | 1.4M | 140k | 0.7M | | | |
| REVERIE[140] | Image, text | 10318 panoramas | 132 | | 4140 | 21702 | 86 |

Table 6 Long Horizon Task Planning Datasets

| Dataset Name | Description | Data Format | Number of Tasks | Task Descriptions | Task Duration Range | Dataset Size (GB) |
|---|---|---|---|---|---|---|
| TaPA[143] | Grounded planning with physical scene constraint | Image, text | | Daily tasks | | 15k instructions |
| RoboCook[151] | Complex long-horizon soft body manipulation tasks | Point clouds | | Complex long-horizon soft body manipulation tasks | | 19.73GB |
| RLBench[263] | An ambitious large-scale benchmark and learning environment | Image, depth, joint angles, velocities and forces | 100 | Manipulation tasks | | |
| PackIt[334] | A dataset of hard packing problems | Point cloud | | The geometry of objects and plan actions for manipulating | | 7.7GB |
| SheetCopilot[335] | A dataset as a foundation for assessing the spreadsheet control capabilities | Text | 221 | Spreadsheet control tasks | | |
| Kinetics[336] | A dataset for inverse kinematics of robotic manipulators | Coordinates of the robotic manipulator end effector | | Inverse kinematics | | 15k data |
| HMDB[337] | A large human motion database | Video | | Human action tasks | | 2GB |
| UCF[338] | An action recognition data set of realistic action videos | Video | 101 | Action recognition tasks | | 6.5GB |
| CALVIN[339] | A benchmark to learn long-horizon language-conditioned tasks | Image, depth, tactile image | 34 | Language-conditioned continuous control tasks | | 1.3GB |
| IKEA assembly dataset[340] | A multi-modal and multi-view video dataset of assembly tasks | 3 RGB views, one depth stream, atomic actions, human poses, object segments, object tracking, and extrinsic camera calibration | | Assembly tasks | | 406GB |
| Epic-Kitchens[341] | A large-scale dataset in first-person (egocentric) vision | Image, audio | 700 | Daily activities in the kitchen | | 1.1TB |
| TUM Kitchen DataSet[342] | A dataset of everyday manipulation activities for motion tracking and action recognition | Image, motion capture data, RFID tag readings | | Action recognition tasks | | |
| MPII Cooking Activities Dataset[343] | A dataset which distinguishes 65 fine-grained activities | Image, video | 65 | Cooking activities | | 141.7GB |
| Coin-Dataset[344] | A large-scale dataset for comprehensive instruction video analysis | Video | 180 | Video analysis tasks | 12 domains as: nursing & caring, vehicles, leisure & performance, gadgets, electric appliances, household items, science & craft, plants & fruits, snacks & drinks dishes, sports, and housework | 476 hours video |
| Youcook2[345] | A large task-oriented, instructional video dataset | Video | 2000 | Cooking tasks | | 35.1GB |
| Kitchen(CMU)[346] | A dataset contains multimodal measures of the human activity of subjects performing the tasks involved in cooking and food preparation. | Video, audio, IMU | 5 | The human activity of subjects performing the tasks involved in cooking and food preparation | | |
| MMACT[347] | A large-scale dataset for cross modal learning on human action understanding | RGB, Keypoints, Acceleration, Gyroscope, Orientation, Wi-Fi, Pressure | 37 | 3 major groups: complex actions, simple action, desk actions | | 40K+ instances |
| FLOBOT Perception[348] | A real-world dataset for robotic cleaning in public spaces | Image, depth, LiDAR | 3(scenes) | Robotic cleaning tasks | | 12.5GB |
| Fusion360Gallery Dataset | A dataset contains rich 2D and 3D geometry data derived from parametric CAD models | B-Rep, mesh, and point cloud | | Design tasks | | |

Table 7 Reasoning Datasets

| Dataset Name | Description | Publisher | Domain/Application | Data Type | Task Type | Size (Number of Samples) |
|---|---|---|---|---|---|---|
| InfLevel[349] | A benchmark designed to evaluate AI systems' understanding of core physical principles | AI2 | Evaluate AI systems' understanding of core physical principles | Video | Manipulation | 80k videos |
| Socratis[214] | A benchmark of diverse open-ended emotional reactions to image-caption pairs | Boston University | Predict human emotion at an image and caption pair | Image, text | Reasoning | 18,378 annotations |
| RoboTurk[258] | A crowdsourced teleoperated robot manipulation dataset | Stanford University | Challenging manipulation tasks | Video | Manipulation | 2144 demonstrations |
| Rainbow[350] | A universal commonsense reasoning benchmark spanning both social and physical common sense | AI2 | Commonsense reasoning tasks | Question-answering | Reasoning | |
| PhysObjects[207] | An object-centric dataset of common household objects | Stanford University etc. | Help VLMs understand the physical concepts (e.g., material, fragility) of common objects | Image, text | Manipulation | 39.6K crowd-sourced and 417K annotations |
| AGENT[351] | A benchmark for core psychological reasoning | MIT etc. | Core psychological reasoning | RGB-D frames, segmentation maps, GT states | Reasoning | 3360 trials |
| Winograd Schema Challenge[352] | A dataset for commonsense | University of Toronto | Commonsense reasoning | Text | Multiple Choice Questions | 150 schemas |
| | reasoning and AI. | | | | | |
| SWAG[353] (Situation With Adversarial Generations) | A dataset for grounded commonsense inference. | University of Washington | Commonsense reasoning | Text | Multiple Choice Questions | 113k examples |
| CLEVR (Compositional Language and Elementary Visual Reasoning)[354] | A dataset for visual reasoning and logic. | Stanford University | Visual reasoning | Images, Text | Multiple Choice Questions | 100k images, 100k questions |
| DROP[355] | A dataset for reading comprehension and reasoning. | AI2 | Reading comprehension | Text | Question Answering | 96k questions |
| SQuAD(The Stanford Question Answering Dataset)[356] | A reading comprehension dataset | Stanford University | Reading comprehension | Text | Question Answering | 150k questions |
| ARC[357] (AI2 Reasoning Challenge) | A set of science questions that require various forms of reasoning. | AI2 | Science reasoning | Text | Multiple Choice Questions | 7k questions |
| HellaSwag[358] | A dataset designed for commonsense reasoning with unconventional questions. | AI2 | Commonsense reasoning | Text | Multiple Choice Questions | 70k questions |
| QuAC[359] (Question Answering in Context) | A dataset for reasoning about questions posed in a dialogue context. | University of Washington | Dialog-based reasoning | Text | Question Answering | 14k dialogues |
| OBQA[360] (OpenBookQA) | A dataset for open-domain question answering that requires reasoning over a large collection of texts. | AI2 and University of Washington | Question answering | Text | Multiple Choice Questions | 6k questions |